\documentclass{article}
\usepackage{preprint}
\usepackage{lmodern}
\usepackage[twoside,letterpaper,includeheadfoot,%
layoutsize={8.125in,10.875in},%
layouthoffset=0.1875in,%
layoutvoffset=0.0625in,%
left=1.0in,%
right=1.0in,%
top=45pt,%
bottom=45pt,%
headheight=0pt,
headsep=10pt,%
footskip=25pt]{geometry}

\usepackage[utf8]{inputenc} 
\usepackage[T1]{fontenc}    
\usepackage{hyperref}       
\usepackage{url}            
\usepackage{booktabs}       
\usepackage{amsfonts}       
\usepackage{nicefrac}       
\usepackage{microtype}      
\usepackage{graphicx}
\usepackage[authoryear]{natbib}
\usepackage{wrapfig}
\usepackage{color}

\title{\LARGE \textbf{Robustness properties of Facebook's ResNeXt WSL models}}

\author{ \large \textbf{Emin Orhan} \\
		\normalsize \texttt{eo41@nyu.edu} \\
		\normalsize New York University
}

\begin{document}

\maketitle

\begin{abstract}
We investigate the robustness properties of ResNeXt class image recognition models trained with billion scale weakly supervised data (ResNeXt WSL models). These models, recently made public by Facebook AI, were trained with $\sim$1B images from Instagram and fine-tuned on ImageNet. We show that these models display an unprecedented degree of robustness against common image corruptions and perturbations, as measured by the ImageNet-C and ImageNet-P benchmarks. They also achieve substantially improved accuracies on the recently introduced ``natural adversarial examples'' benchmark (ImageNet-A). The largest of the released models, in particular, achieves state-of-the-art results on ImageNet-C, ImageNet-P, and ImageNet-A by a large margin. The gains on ImageNet-C, ImageNet-P, and ImageNet-A far outpace the gains on ImageNet validation accuracy, suggesting the former as more useful benchmarks to measure further progress in image recognition. Remarkably, the ResNeXt WSL models even achieve a limited degree of adversarial robustness against state-of-the-art white-box attacks (10-step PGD attacks). However, in contrast to adversarially trained models, the robustness of the ResNeXt WSL models rapidly declines with the number of PGD steps, suggesting that these models do not achieve genuine adversarial robustness. Visualization of the learned features also confirms this conclusion. Finally, we show that although the ResNeXt WSL models are more shape-biased than comparable ImageNet-trained models in a shape-texture cue conflict experiment, they still remain much more texture-biased than humans, suggesting that they share some of the underlying characteristics of ImageNet-trained models that make this benchmark challenging.
\end{abstract}

\section{Introduction}
Facebook AI recently released ResNeXt class image recognition models trained with $\sim$1B images from Instagram using weak supervision (i.e. with noisy labels) and fine-tuned on ImageNet. To our knowledge, these models are the only publicly available models trained with such large scale data. These models can help us address some important and interesting questions about the relationship between training data size and the out-of-sample generalization behavior of image recognition models trained in standard classification tasks. For example: does more training data make the learned representations more robust against common image corruptions and perturbations? Does it make them more robust against adversarial attacks? Does it reduce or even eliminate some of the quirky behavior of ImageNet-trained models, such as their sensitivity to background cues, their heavy reliance on local textural information, and their surprising inability to integrate information more globally across an image \citep{geirhos2019}? In this paper, we address these questions.\footnote{Code and all simulation results are available at: \url{https://github.com/eminorhan/resnext-wsl}}

Intuitively, we expect that training with more data should in general increase the robustness of a model, because more data constrain the behavior of the model more strongly. But the scaling of different robustness measures with training data size is an open empirical question. We find that the models trained with billion scale data are substantially more robust than ImageNet-trained models on common image corruptions and perturbations, achieving state-of-the-art results on both ImageNet-C and ImageNet-P benchmarks \citep{hendrycks2019} by a large margin. They also achieve substantially improved accuracies on the recently introduced ``natural adversarial examples'' measuring out-of-sample generalization performance on adversarially selected natural images \citep{hendrycks2019b}. These models even achieve a limited degree of robustness against white-box adversarial attacks. However, it remains relatively easy to generate adversarial examples for them, hence they do not achieve true adversarial robustness. They also retain the strong texture bias of ImageNet-trained models, suggesting that these issues are unlikely to be solved by simply increasing the training data size.
\section{Methods}
\subsection{Models}
We consider five different models. The models all belong the ResNeXt family \citep{xie2017}. The first one was trained on ImageNet; the remaining four models (WSL models) were trained on $\sim$1B images from Instagram using weak supervision and then fine-tuned on ImageNet (please see \citet{mahajan2018} for further details about training).\footnote{The WSL models can be accessed from: \url{https://pytorch.org/hub/facebookresearch_WSL-Images_resnext/}}

\textbf{\texttt{resnext101\_32x8d}.} This is an ImageNet-trained ResNeXt-101 model with cardinality 32 and a bottleneck width of 8 (please see \citet{xie2017} for a description of the different architectural dimensions). We use the implementation of this model in \texttt{torchvision.models} (0.3).

\textbf{\texttt{resnext101\_32x8d\_wsl}.} This model has the same architecture as the previous one, but was trained on Instagram images. The comparison between this model and the previous one is important, because any difference between these models is due to the difference in training data.

\textbf{\texttt{resnext101\_32x16d\_wsl}.} Instagram-trained ResNeXt-101 model with cardinality 32 and a bottleneck width of 16.

\textbf{\texttt{resnext101\_32x32d\_wsl}.} Instagram-trained ResNeXt-101 model with cardinality 32 and a bottleneck width of 32.

\textbf{\texttt{resnext101\_32x48d\_wsl}.} Instagram-trained ResNeXt-101 model with cardinality 32 and a bottleneck width of 48. With $\sim$829M parameters, this is the largest WSL model released by Facebook AI.

\subsection{Measuring the robustness against common image corruptions and perturbations}
We measured the robustness of the models against common natural corruptions and perturbations using the recently introduced ImageNet-C and ImageNet-P benchmarks \citep{hendrycks2019}. We give a brief description of these benchmarks below and refer the reader to \citet{hendrycks2019} for further details.

ImageNet-C was designed to measure the robustness of classifiers against common image corruptions and contains 15 different corruption types\footnote{Gaussian, shot, and impulse noise; defocus, glass, motion, and zoom blur; snow, frost, fog, and brightness corruptions; contrast, elasticity, pixelation, and JPEG compression.} applied to each ImageNet validation image at 5 different severity levels. 

The robustness performance on ImageNet-C is measured by the mean corruption error (\textit{mCE}), which is defined as follows. For each corruption type $c$, the classification error of the model is averaged over different severity levels $s$ and then divided by the average classification error of a reference classifier (AlexNet): i.e. $CE_c \equiv \langle E_{s,c} \rangle_s / \langle E_{s,c}^{\mathrm{AlexNet}} \rangle_s$. The mean corruption error is then obtained by averaging over the corruption types: $mCE \equiv \langle CE_c \rangle_c$. We also calculate a relative \textit{mCE} (\textit{rel. mCE}) score by subtracting the clean classification error of the classifiers from the corruption errors: i.e. $rel. \; CE_c \equiv \langle E_{s,c} - E_{\mathrm{clean}}\rangle_s / \langle E_{s,c}^{\mathrm{AlexNet}} - E_{\mathrm{clean}}^{\mathrm{AlexNet}} \rangle_s$ and then averaging over different corruption types as before.

ImageNet-P was designed to measure the stability of a model's predictions as the input image undergoes a continuous sequence of transformations. ImageNet-P contains 10 common perturbation types\footnote{Gaussian noise, shot noise, motion blur, zoom blur, brightness, snow, translation, rotation, tilt, scale perturbations.} applied to each ImageNet validation image in a temporal sequence. Each sequence contains more than 30 images.

The robustness performance on ImageNet-P is measured by the mean flip rate (\textit{mFR}) and the mean top-5 distance (\textit{mT5D}) metrics. The mean flip rate is calculated by first computing the flip probability of the model's predictions for consecutive frames for each perturbation $p$, $FP_p \equiv \Pr_{x\sim p}(f(x_t) \neq f(x_{t-1}))$, normalizing by the AlexNet flip probability for the same perturbation, and then averaging over different perturbations: $mFR \equiv \langle FP_p / FP_p^{\mathrm{AlexNet}} \rangle_p$. The flip probability is computed somewhat differently for the noise perturbations, where the consecutive frames are not temporally related. We refer the reader to \citet{hendrycks2019} for more details. The mean top-5 distance is defined similarly, but instead of the stability of the model's top-1 prediction for consecutive frames, it measures the stability of the top-5 predictions. We again refer the reader to \citet{hendrycks2019} for further details.

\subsection{Measuring the adversarial robustness}
We considered both black-box and white-box attacks to measure the robustness of the models against adversarial perturbations. Attacks were carried out with the state-of-the-art projected gradient descent (PGD) algorithm using the Foolbox implementation \citep{rauber2017}. 

\textbf{Black-box attacks.} In the black-box setting, we ran attacks against the \texttt{resnext50\_32x4d} model in \texttt{torchvision.models}. Note that this model is different from the five models considered in this paper. We set the number of PGD steps to $10$ and the step size to $2/225$. We varied the total perturbation size of the attack $\epsilon$, defined as the $l_\infty$-norm of the perturbation divided by the $l_\infty$-norm of the clean image: $\epsilon \equiv ||\mathbf{x}_{\mathrm{adv}} - \mathbf{x}||_\infty / ||\mathbf{x}||_\infty$, from $0.01$ to $0.1$. These attacks against the \texttt{resnext50\_32x4d} model were highly successful, yielding below $10\%$ top-1 accuracy even for the lowest perturbation size $\epsilon=0.01$. We then tested the generated adversarial images with the five ResNeXt-101 models considered in this paper. 

\textbf{White-box attacks.} In the white-box setting, attacks were run directly against the ResNeXt-101 models. Attack parameters were identical to those described in the previous paragraph. However, since using only a small number of PGD steps can lead to a significant overestimation of the robustness of a model against white-box adversarial attacks \citep{engstrom2018}, we also ran stronger white-box attacks with up to $50$ PGD steps (fixing the total perturbation size to $\epsilon=0.06$ for these attacks). 

\subsection{Visualization of the learned features}
\citet{engstrom2019} recently showed that models trained with robust optimization learn fundamentally different features from models trained in the standard way (through minimization of training loss). In particular, they show that the learned features in robust models are much more meaningful and well-aligned with human perception than the learned features in non-robust models. Here, we use this idea to test whether the learned features in ResNeXt WSL models show this signature of robustness. Following \citet{engstrom2019}, we do this by starting from a seed image and finding an image that maximizes the activation of a particular unit in the penultimate layer of the network. \citet{engstrom2019} show that the resulting ``maximizing images'' are much more meaningful and much less sensitive to the initial seed image in robust models than in non-robust models. To find these ``maximizing images'', we removed the final softmax layer from the network and used the PGD algorithm to maximize different units in the penultimate layer of the network. We used the Foolbox implementation \texttt{ProjectedGradientDescentAttack} with the \texttt{TargetClassProbability} criterion set to a large value ($1-10^{-6}$) for the corresponding unit. Note that this is slightly different from the way maximizing images were computed in \citet{engstrom2019}.

\subsection{Measuring the shape bias}
To test whether the Instagram-trained ResNeXt WSL models share the characteristic texture bias displayed by ImageNet-trained deep neural networks, we used the shape-texture cue conflict stimuli created by \citet{geirhos2019}. These are 1201 images created with a neural style transfer algorithm to look locally like an image from a given category (texture content), but globally like an image from a different category (shape content). Therefore, these images are ideal for testing a model's relative sensitivity to local texture information vs. global shape information. \citet{geirhos2019} showed that ImageNet-trained deep neural networks rely much more heavily on local texture information than on global shape information in making their predictions. This was found to be in stark contrast to humans who are sensitive to both local and global information, but rely almost exclusively on global shape in making classification judgments. We used the same evaluation procedure as \citet{geirhos2019} to measure the shape/texture biases of the models. Briefly, the images were generated from 16 distinct super-categories in ImageNet. In evaluating the predictions of the models and hence measuring their shape/texture biases, only ImageNet classes belonging to these 16 super-categories were considered, the remaining classes being zeroed out. We refer the reader to \citet{geirhos2019} for further details about the stimulus generation and model evaluation methods.

\subsection{Measuring the robustness against ``natural adversarial examples''}
Finally, we measured the performance of the ResNeXt WSL models on the recently introduced ImageNet-A dataset \citep{hendrycks2019b}. This curated dataset consists of 7500 natural, unmodified ImageNet-like images for which a standard ImageNet-trained classifier yields incorrect predictions with low confidence in the correct class (less than 15\%). These ``natural adversarial examples'' were also selected to display a diverse range of confusions between different classes. \citet{hendrycks2019b} argue that misclassifications on the dataset result from a diverse range of underlying causes, such as over-reliance on texture, color, or background cues, sensitivity to image distortions or perturbations, tendency to over-generalize etc. 

The images in ImageNet-A belong to a subset of 200 classes among the 1000 ImageNet-1K classes. Accuracies are measured on this 200-class subset only (outputs corresponding to the remaining classes being effectively zeroed out). \citet{hendrycks2019b} also introduce two uncertainty metrics to quantify the confidence miscalibration of models: the RMS calibration error (RMS-CE) and the area under the response rate accuracy curve (AURRA). We refer the reader to \citet{hendrycks2019b} for a detailed description of how these metrics are calculated.

\section{Results}
\subsection{ResNeXt WSL models are highly robust against natural perturbations}
ImageNet-C and ImageNet-P robustness scores are reported in Table~\ref{inc_inp_table}. The ResNeXt WSL models outperform the ImageNet-trained \texttt{resnext101\_32x8d} model on all metrics. The largest WSL model \texttt{resnext101\_32x48d\_wsl}, in particular, achieves state-of-the-art results on all metrics by a large margin. The robustness gains achieved by the WSL models over the ImageNet-trained \texttt{resnext101\_32x8d} model are significantly larger than the the gains achieved on ImageNet validation accuracy (ImageNet validation accuracies are reported in Table~\ref{adv_table} under the ``Clean'' column). This suggests that robustness on ImageNet-C and ImageNet-P may be a more meaningful metric than ImageNet validation accuracy in evaluating future improvements in image recognition models. 

\begin{table}
	\caption{ImageNet-C and ImageNet-P robustness scores. For all metrics, lower values indicate more robust models. See the \textit{Methods} section for a detailed description of the robustness metrics. The best previously reported robustness scores are included in the bottom two rows of the table for reference.}
	\vspace{0.25cm}
	\label{inc_inp_table}
	\centering
	\renewcommand{\arraystretch}{1.25}
\begin{tabular}{lcccc}
	\toprule
	& \multicolumn{2}{c}{ImageNet-C} & \multicolumn{2}{c}{ImageNet-P} \\
	Model                  			 & \textit{mCE} & \textit{rel. mCE} & \textit{mT5D} & \textit{mFR}    \\
	\midrule
	\texttt{resnext101\_32x8d}       &  66.6       &   91.1           &     70.5       &     46.1         \\
	\texttt{resnext101\_32x8d\_wsl}  &  51.7       &   65.6           &     58.2       &     38.8         \\
	\texttt{resnext101\_32x16d\_wsl} &  48.8       &   65.0           &     53.7       &     32.7         \\
	\texttt{resnext101\_32x32d\_wsl} &  47.0       &   63.8           &     53.9       &     29.7         \\
	\texttt{resnext101\_32x48d\_wsl} & \textbf{45.7} & \textbf{61.8} & \textbf{52.9} & \textbf{27.8}      \\
	\midrule
	Patch Gaussian (ResNet-200) \citep{lopes2019} & 60.4 & 75.7 &  -- & --          \\
	\texttt{resnext101\_64x4d} \citep{hendrycks2019} & 62.2 & 80.1 & 65.9 & 43.2 \\
	\bottomrule
\end{tabular}
\end{table}

\subsection{ResNeXt WSL models achieve limited robustness against white-box adversarial attacks}
\begin{figure}
	\includegraphics[width=1\textwidth, trim=0mm 0mm 0mm 0mm, clip]{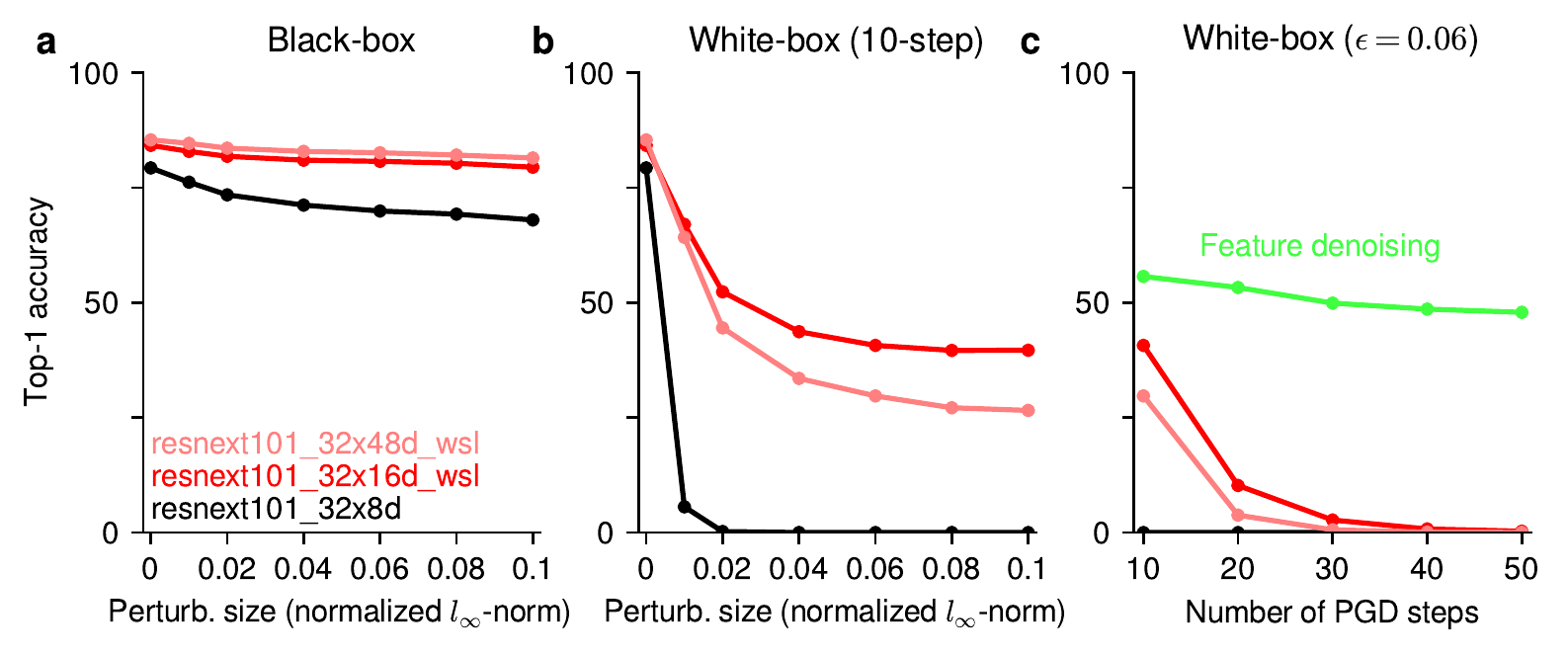}
	\caption{(\textbf{a-b}) Top-1 accuracy under black-box and white-box adversarial attacks, respectively, as a function of the total perturbation size. The attacks were carried out with a 10-step PGD algorithm. The $0$ perturbation size corresponds to the clean images. (\textbf{c}) Top-1 accuracy under white-box adversarial attacks as a function of the number of PGD steps, fixing the total perturbation size to $\epsilon=0.06$. For comparison, the performance of a state of the art, adversarially trained ResNet-152 model \citep{xie2018} is also included in the figure (labeled {\color{green}feature denoising}). Unlike the ResNeXt WSL models, the robustness of the adversarially trained model remains much more stable as the number of PGD steps is increased.} \label{adversarial_fig}
\end{figure}

The robustness of the models against black-box and white-box adversarial attacks is shown in Table~\ref{adv_table} and in Figure~\ref{adversarial_fig}. The ResNeXt WSL models achieve significantly better black-box adversarial accuracy compared to the ImageNet-trained \texttt{resnext101\_32x8d} model. Even more impressively, however, the WSL models also achieve a significant amount of robustness against 10-step white-box PGD attacks. Note that a 10-step PGD attack is strong enough to yield close to 0\% accuracy on the ImageNet-trained \texttt{resnext101\_32x8d} model for a standard perturbation size of $\epsilon=0.06$. By comparison, the best WSL model (\texttt{resnext101\_32x16d\_wsl}) yields an accuracy of 40.7\% in the same condition. In fact, this level of robustness is better than that achieved by some previous adversarial training methods (e.g. ALP, see Table~\ref{adv_table}). This is surprising given that the WSL models were not explicitly trained to be adversarially robust and suggests that simply training models with more data can automatically improve adversarial robustness.

\citet{gilmer2019} recently argued that adversarial vulnerability and sensitivity to more common image corruptions and perturbations are two sides of the same underlying phenomenon, namely sensitivity to perturbations in general. According to their perspective, adversarial non-robustness simply arises as the worst-case manifestation of this general sensitivity to perturbations, whereas sensitivity to more common image corruptions and perturbations is the average-case manifestation of the same. This implies that robustness gains in one should, in general, accompany robustness gains in the other measure. Given the large gains in robustness to common image corruptions and perturbations and the concomitant gains in adversarial robustness observed in the WSL models, our results are consistent with this prediction of \citet{gilmer2019}.

The adversarial robustness of the WSL models, however, declined rapidly when we increased the number of PGD iterations up to 50 steps, fixing the total perturbation  size to $\epsilon=0.06$ (Figure~\ref{adversarial_fig}c). This is in contrast to the robustness achieved by a state-of-the-art adversarially-trained model (feature denoising with a ResNet-152 backbone, shown in green in Figure~\ref{adversarial_fig}c), which remains much more stable as the number of PGD iterations is increased. This result suggests that the ResNeXt WSL models do not achieve true adversarial robustness.

\begin{table}
	\caption{Robustness against black-box and white-box adversarial perturbations (top-1 accuracy). For comparison, top-1 accuracies of two adversarially trained ImageNet models (ALP and feature denoising) are included in the bottom two rows of the table.}
	\vspace{0.25cm}
	\label{adv_table}
	\centering
	\renewcommand{\arraystretch}{1.25}
	\begin{tabular}{lcccc}
		\toprule
		Model & Clean & Black-box         & White-box                  & White-box           \\
			  &       & {\scriptsize($\epsilon=0.06$)} & {\scriptsize(10-step, $\epsilon=0.06$)} & {\scriptsize(50-step, $\epsilon=0.06$)} \\ 
		\midrule
		\texttt{resnext101\_32x8d}       &  79.3  &  69.9  &  0.0   &  0.0         \\
		\texttt{resnext101\_32x8d\_wsl}  &  82.2  &  78.3  &  34.4  &  0.1         \\
		\texttt{resnext101\_32x16d\_wsl} &  84.2  &  80.7  &  \textbf{40.7} & \textbf{0.3}         \\
		\texttt{resnext101\_32x32d\_wsl} &  85.1  &  82.0  &  30.6  &  0.1         \\
		\texttt{resnext101\_32x48d\_wsl} &  \textbf{85.4}  &  \textbf{82.6} &  29.7  &  0.0         \\
		\midrule
		ALP (InceptionV3) \citep{kannan2018} & 72 & -- & 27.9 & -- \\
		Denoising (ResNet-152) \citep{xie2018} & 65.3 & -- & 55.7 & 47.9 \\
		\bottomrule
	\end{tabular}
\end{table}

\subsection{The learned features in ResNeXt WSL models show signatures of adversarial non-robustness}
\begin{figure}
	\includegraphics[width=1\textwidth, trim=0mm 0mm 0mm 0mm, clip]{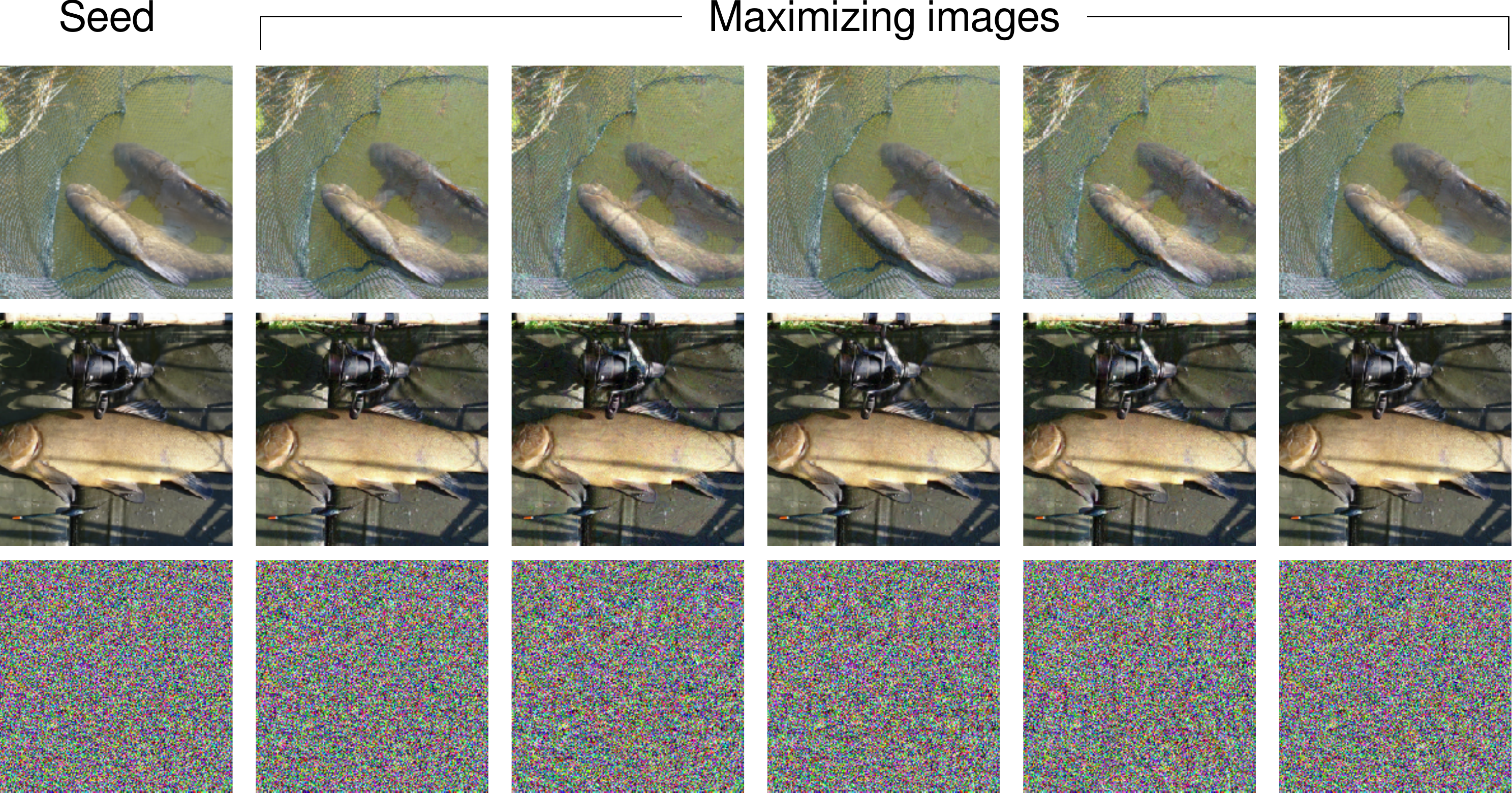}
	\caption{Maximizing images for the \texttt{resnext101\_32x8d} model. The leftmost column shows three seed images; each of the remaining five columns shows the maximizing image for a particular unit of the penultimate layer in the network, starting from the corresponding seed image. The maximizing images are perceptually very similar to the corresponding seed images. \citet{engstrom2019} show that such extreme sensitivity to initial seeds is a sign of the non-robustness of the learned features.} \label{feats_fig}
\end{figure}

\begin{figure}
	\includegraphics[width=1\textwidth, trim=0mm 0mm 0mm 0mm, clip]{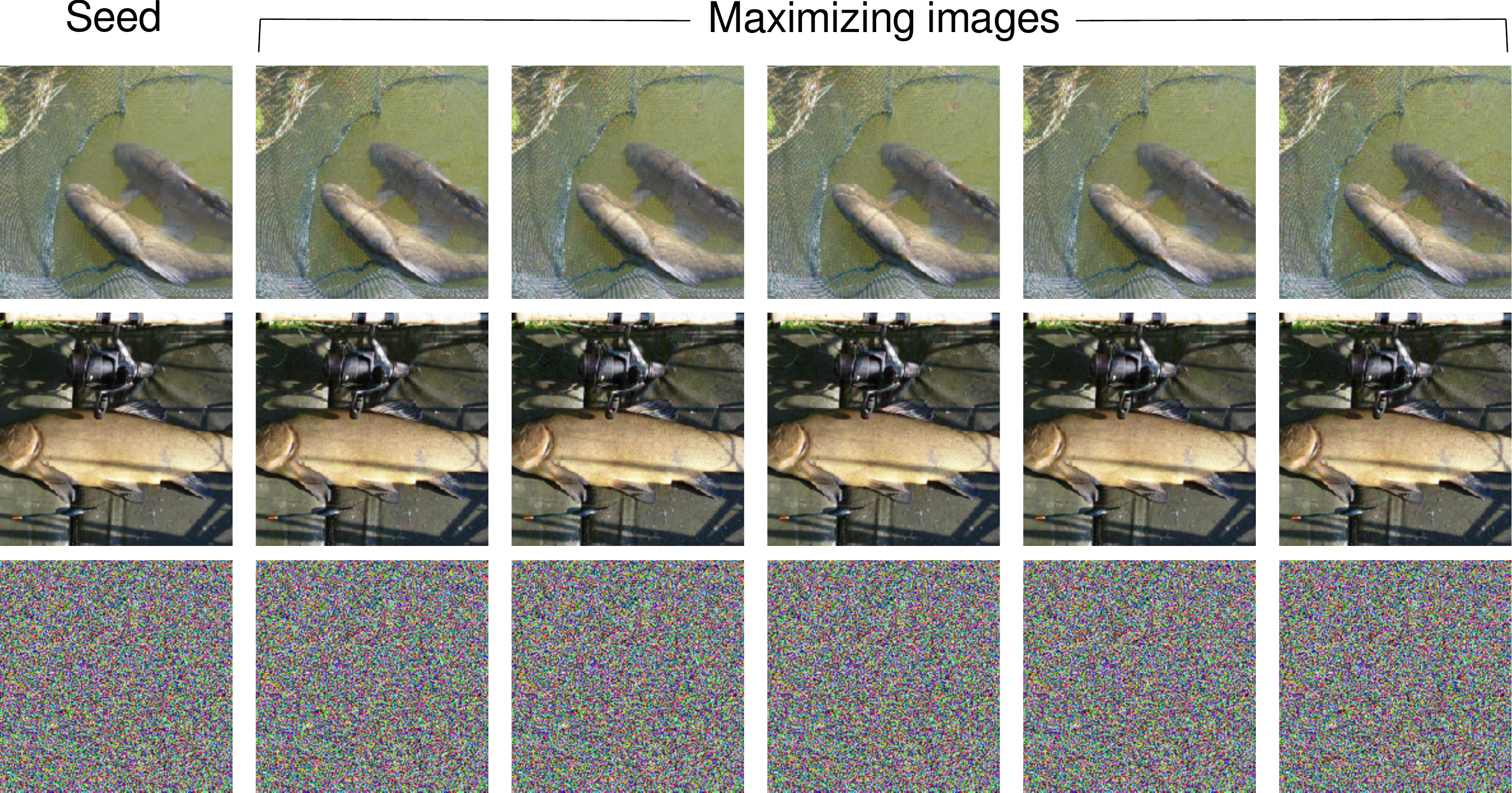}
	\caption{Maximizing images for the \texttt{resnext101\_32x48d\_wsl} model. Similar to Figure~\ref{feats_fig}.} \label{feats_wsl_fig}
\end{figure}

Maximizing images for the ImageNet-trained \texttt{resnext101\_32x8d} and the Instagram-trained \texttt{resnext101\_32x48d\_wsl} models are shown in Figures~\ref{feats_fig} and \ref{feats_wsl_fig}, respectively, together with the seed images used in optimization. Both models produce qualitatively similar maximizing images. The maximizing images essentially look like adversarial examples. Perceptually, the maximizing images for different units are almost identical to each other and to the seed image. \citet{engstrom2019} recently showed that these properties are signatures of adversarially non-robust models (robust models yield perceptually meaningful and heterogeneous maximizing images for different units and the maximizing images are much less dependent on the seed image). This result supports our conclusion from the previous subsection that the ResNeXt WSL models do not achieve genuine adversarial robustness.

\subsection{Billion scale training increases the shape bias, but the resulting models are still far more texture-biased than humans}
\begin{figure}
	\includegraphics[width=1\textwidth, trim=0mm 0mm 0mm 0mm, clip]{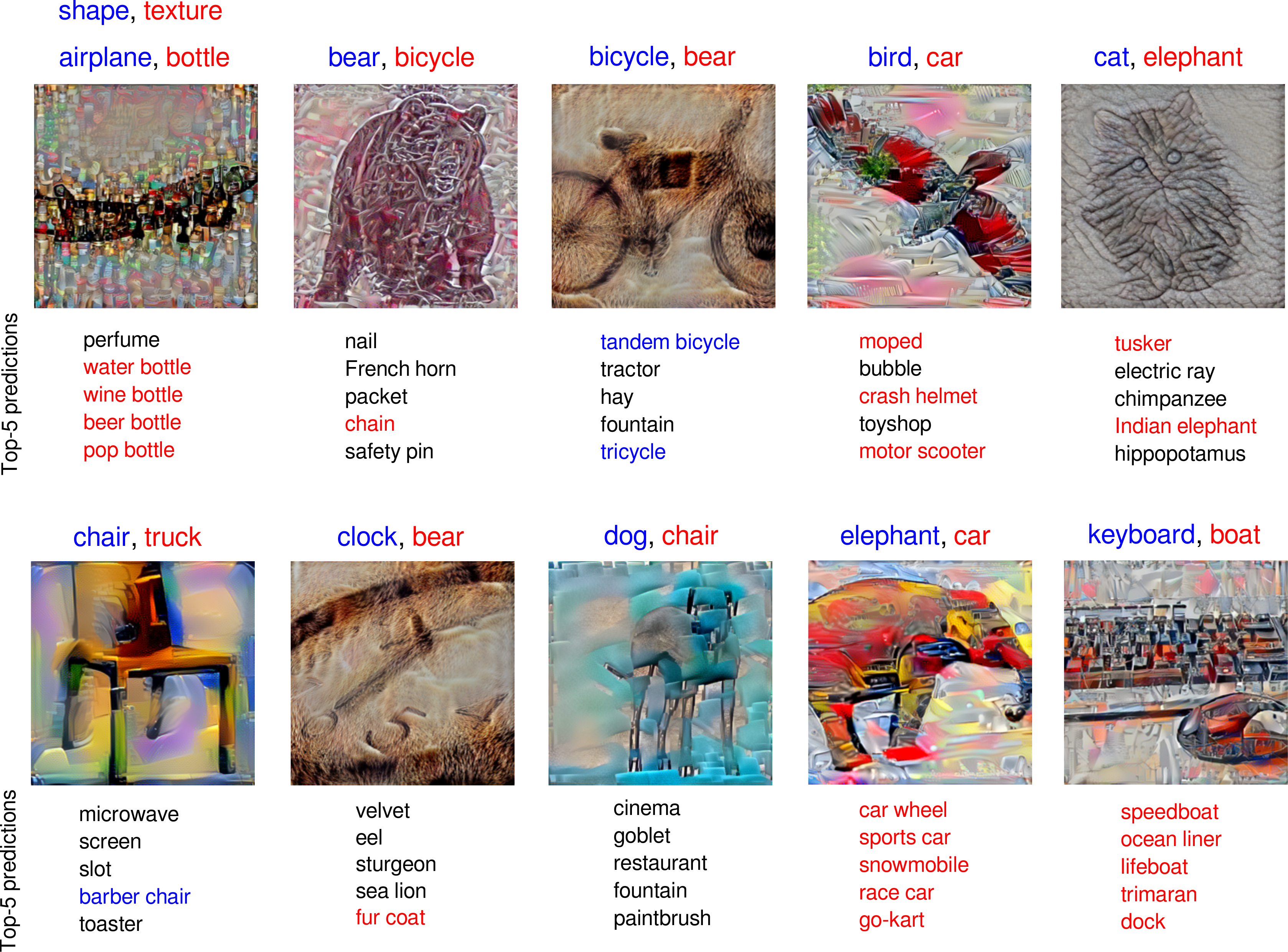}
	\caption{Shape-texture cue conflict images from \citet{geirhos2019}. The {\color{blue}shape} and {\color{red}texture} categories for each image are indicated above the image. The unrestricted top 5 predictions of the \texttt{resnext101\_32x48d\_wsl} model for each image are shown below the image, with approximate shape matches highlighted in blue and approximate texture matches highlighted in red. The shape and texture matches were evaluated manually by the author for this figure only. As described in detail in the \textit{Methods} section, evaluation of the shape bias scores reported in Table~\ref{shape_bias_table} followed a somewhat different procedure and was done in the same way as in \citet{geirhos2019}.} \label{shape_bias_fig}
\end{figure}

The shape biases of different models are reported in Table~\ref{shape_bias_table}. Although billion scale training with Instagram images increases the shape biases of the WSL models compared to ImageNet-trained ResNet and ResNeXt models, the resulting models are still far more texture-biased than humans. Some example shape-texture cue-conflict stimuli are shown in Figure~\ref{shape_bias_fig}, together with the top 5 predictions of the \texttt{resnext101\_32x48d\_wsl} model. 

This result is expected if the statistical regularities enabling high classification performance in Instagram are similar to those observed in ImageNet and standard image recognition models have an inductive bias for exploiting local textural regularities over more global shape-based regularities \citep{brendel2019}.

\begin{table}
	\caption{The shape biases of different models.}
	\vspace{0.25cm}
	\label{shape_bias_table}
	\centering
	\renewcommand{\arraystretch}{1.25}
	\begin{tabular}{lc}
		\toprule
		Model & Shape bias \\
		\midrule
		\texttt{resnext101\_32x8d}       &  25.9     \\
		\texttt{resnext101\_32x8d\_wsl}  &  39.1     \\
		\texttt{resnext101\_32x16d\_wsl} &  42.7     \\
		\texttt{resnext101\_32x32d\_wsl} &  40.6     \\
		\texttt{resnext101\_32x48d\_wsl} &  42.8     \\
		\midrule
		ResNet-50 \citep{geirhos2019} &  22.1     	\\
		Shape-ResNet-50 \citep{geirhos2019} &  81   \\
		Humans \citep{geirhos2019} &  95.9    		\\
		\bottomrule
	\end{tabular}
\end{table}

\subsection{Billion scale training substantially improves the accuracy on ``natural adversarial examples''}
Table~\ref{imageneta_table} shows the top-1 accuracies and confidence miscalibration scores of the models on ImageNet-A. The ImageNet-trained \texttt{resnext101\_32x8d} model achieves a top-1 accuracy of 10.2\%, demonstrating the difficulty of this benchmark for standard ImageNet-trained models. Note that this is in sharp contrast to the performance of ImageNet-trained models on the ImageNetV2 dataset \citep{recht2019}, where despite a significant 11--14\% absolute drop in accuracy, the models remain relatively high-performing. The difference between ImageNetV2 and ImageNet-A is that the images in ImageNet-A were explicitly chosen to be hard for an ImageNet-trained classifier (thus the name ``natural \textit{adversarial} examples''), whereas the images in ImageNetV2 were selected in a way that matched as closely as possible the way the original ImageNet validation set was selected.

The Instagram-trained WSL models achieve much better calibration scores and accuracies on ImageNet-A: in particular, the largest WSL model achieves a top-1 accuracy of 61.0\%; however, even for this model, there remains a substantial $\sim$25\% absolute gap between the accuracy on the ImageNet validation set vs. the accuracy on ImageNet-A, suggesting considerable room for further progress on this benchmark.

\begin{table}
	\caption{Top-1 accuracy and confidence miscalibration scores on ImageNet-A. Note that lower RMS-CE and higher AURRA values indicate better calibrated models. On all three metrics, the largest WSL model performs the best.}
	\vspace{0.25cm}
	\label{imageneta_table}
	\centering
	\renewcommand{\arraystretch}{1.25}
	\begin{tabular}{lccc}
		\toprule
		Model & Top-1 acc. & RMS-CE & AURRA \\
		\midrule
		\texttt{resnext101\_32x8d}       &  10.2  &  54.5  &  12.3 \\
		\texttt{resnext101\_32x8d\_wsl}  &  45.4  &  26.8  &  66.3 \\
		\texttt{resnext101\_32x16d\_wsl} &  53.1  &  22.8  &  75.0 \\
		\texttt{resnext101\_32x32d\_wsl} &  58.1  &  19.0  &  80.2 \\
		\texttt{resnext101\_32x48d\_wsl} &  \textbf{61.0}  &  \textbf{17.6}  &  \textbf{82.4} \\
		\bottomrule
	\end{tabular}
\end{table}

\section{Discussion}
Our results paint a mixed picture regarding the robustness properties of the ResNeXt WSL models trained with billion scale weakly-supervised data. On the one hand, these models achieve a remarkable degree of robustness against common image corruptions and perturbations and even a limited degree of adversarial robustness despite not having been explicitly trained for adversarial robustness, demonstrating yet another example of the ``unreasonable effectiveness of data'' \citep{halevy2009}. On the other hand, they do not achieve genuine adversarial robustness and they retain some of the quirky behavior of ImageNet-trained models, such as their over-reliance on local texture cues and their seeming inefficiency in integrating information more globally across an image.

Although training with more data can clearly improve the generalization behavior of image recognition models, it is unclear at the moment whether simply scaling up the standard object classification tasks and models to even more data will be sufficient to feasibly achieve genuinely human-like, more general-purpose visual representations: adversarially robust, more shape-based and, in general, better able to handle out-of-sample generalization. It remains a big open question what kinds of tasks and model biases can enable the learning of such robust, general-purpose visual representations. As we continue to deploy machine learning models in more and more challenging, open-ended domains, the need for such robust, general-purpose visual representations will likely increase as well. In the meantime, we can be duly impressed by the performance of current generation large scale vision models trained with large amounts of data on more restricted domains.

\section*{Acknowledgments}
We are very grateful to the authors of \citet{geirhos2019}, \citet{hendrycks2019}, \cite{hendrycks2019b}, \citet{mahajan2018}, \citet{rauber2017} for making their pre-trained models, code, and stimuli publicly available, without which this work would not have been possible.
\bibliography{resnext-wsl}

\begin{thebibliography}{}

\bibitem[Brendel and Bethge, 2019]{brendel2019}
Brendel, W. and Bethge, M. (2019).
\newblock Approximating {CNN}s with bag-of-local-features models works
  surprisingly well on {I}mage{N}et.
\newblock In {\em International Conference on Learning Representations (ICLR)}.

\bibitem[Engstrom et~al., 2018]{engstrom2018}
Engstrom, L., Ilyas, A., and Athalye, A. (2018).
\newblock Evaluating and understanding the robustness of adversarial logit
  pairing.
\newblock {\em arXiv preprint arXiv:1807.10272}.

\bibitem[Engstrom et~al., 2019]{engstrom2019}
Engstrom, L., Ilyas, A., Santurkar, S., Tsipras, D., Tran, B., and Madry, A.
  (2019).
\newblock Learning perceptually-aligned representations via adversarial
  robustness.
\newblock {\em arXiv preprint arXiv:1906.00945}.

\bibitem[Geirhos et~al., 2019]{geirhos2019}
Geirhos, R., Rubisch, P., Michaelis, C., Bethge, M., Wichmann, F.~A., and
  Brendel, W. (2019).
\newblock Image{N}et-trained {CNN}s are biased toward texture; increasing shape
  bias improves accuracy and robustness.
\newblock In {\em International Conference on Learning Representations (ICLR)}.

\bibitem[Gilmer et~al., 2019]{gilmer2019}
Gilmer, J., Ford, N., Carlini, N., and Cubuk, E. (2019).
\newblock Adversarial examples are a natural consequence of test error in
  noise.
\newblock In {\em International Conference on Machine Learning}, pages
  2280--2289.

\bibitem[Halevy et~al., 2009]{halevy2009}
Halevy, A., Norvig, P., and Pereira, F. (2009).
\newblock The unreasonable effectiveness of data.
\newblock {\em IEEE Intelligent Systems}, 24(2):8--12.

\bibitem[Hendrycks and Dietterich, 2019]{hendrycks2019}
Hendrycks, D. and Dietterich, T. (2019).
\newblock Benchmarking neural network robustness to common corruptions and
  perturbations.
\newblock In {\em International Conference on Learning Representations (ICLR)}.

\bibitem[Hendrycks et~al., 2019]{hendrycks2019b}
Hendrycks, D., Zhao, K., Basart, S., Steinhardt, J., and Song, D. (2019).
\newblock Natural adversarial examples.
\newblock {\em arXiv preprint arXiv:1907.07174}.

\bibitem[Kannan et~al., 2018]{kannan2018}
Kannan, H., Kurakin, A., and Goodfellow, I. (2018).
\newblock Adversarial logit pairing.
\newblock {\em arXiv preprint arXiv:1803.06373}.

\bibitem[Lopes et~al., 2019]{lopes2019}
Lopes, R.~G., Yin, D., Poole, B., Gilmer, J., and Cubuk, E.~D. (2019).
\newblock Improving robustness without sacrificing accuracy with patch
  {G}aussian augmentation.
\newblock {\em arXiv preprint arXiv:1906.02611}.

\bibitem[Mahajan et~al., 2018]{mahajan2018}
Mahajan, D., Girshick, R., Ramanathan, V., He, K., Paluri, M., Li, Y.,
  Bharambe, A., and van~der Maaten, L. (2018).
\newblock Exploring the limits of weakly supervised pretraining.
\newblock In {\em Proceedings of the European Conference on Computer Vision
  (ECCV)}, pages 181--196.

\bibitem[Rauber et~al., 2017]{rauber2017}
Rauber, J., Brendel, W., and Bethge, M. (2017).
\newblock Foolbox: A {P}ython toolbox to benchmark the robustness of machine
  learning models.
\newblock {\em arXiv preprint arXiv:1707.04131}.

\bibitem[Recht et~al., 2019]{recht2019}
Recht, B., Roelofs, R., Schmidt, L., and Shankar, V. (2019).
\newblock Do {I}mage{N}et classifiers generalize to {I}mage{N}et?
\newblock {\em arXiv preprint arXiv:1902.10811}.

\bibitem[Xie et~al., 2018]{xie2018}
Xie, C., Wu, Y., van~der Maaten, L., Yuille, A., and He, K. (2018).
\newblock Feature denoising for improving adversarial robustness.
\newblock {\em arXiv preprint arXiv:1812.03411}.

\bibitem[Xie et~al., 2017]{xie2017}
Xie, S., Girshick, R., Doll{\'a}r, P., Tu, Z., and He, K. (2017).
\newblock Aggregated residual transformations for deep neural networks.
\newblock In {\em Proceedings of the IEEE Conference on Computer Vision and
  Pattern Recognition}, pages 1492--1500.

\end{thebibliography}
\bibliographystyle{apalike}

\end{document}